\documentclass[10pt]{article}

\usepackage{jmlr2e}
\usepackage[font={scriptsize}]{subcaption}
\usepackage{graphicx}
\usepackage{amsmath}
\usepackage{xcolor,colortbl}
\usepackage{array}

\newcolumntype{P}[1]{>{\centering\arraybackslash}p{#1}}
\title{TimeNet: Pre-trained deep recurrent neural network for time series classification}

\author{\name Pankaj Malhotra \email malhotra.pankaj@tcs.com\\
	\name Vishnu TV \email vishnu.tv@tcs.com\\
	\name Lovekesh Vig \email Lovekesh.vig@tcs.com\\
	\name Puneet Agarwal \email puneet.a@tcs.com\\
	\name Gautam Shroff \email gautam.shroff@tcs.com\\
	\addr TCS Research, New Delhi, India}

\begin{document}
\maketitle

\begin{abstract}
In the spirit of the tremendous success of deep Convolutional Neural Networks as generic feature extractors from images, we propose \textit{Timenet}: a multilayered recurrent neural network (RNN) trained in an unsupervised manner 
to extract features from time series\footnote{This is an extended version of the work published as a conference paper in the Proceedings of 25th European Symposium on Artificial Neural Networks, Computational Intelligence and Machine Learning.\\Copyright $\copyright$ 2017 Tata Consultancy Services Ltd.}. 
Fixed-dimensional vector representations or embeddings of variable-length sentences have been shown to be useful for a variety of document classification tasks. Timenet is the encoder network of an auto-encoder based on sequence-to-sequence models that transforms varying length time series to fixed-dimensional vector representations.
Once Timenet is trained on diverse sets of time series, it can then be used as a \textit{generic off-the-shelf feature extractor for time series}. 
We train Timenet on time series from 24 datasets belonging to various domains from the UCR Time Series Classification Archive, and then evaluate embeddings from Timenet for classification on 30 other datasets not used for training the Timenet. We observe that a classifier learnt over the embeddings obtained from a pre-trained Timenet yields significantly better performance compared to (i) a classifier learnt over the embeddings obtained from the encoder network of a domain-specific auto-encoder, as well as (ii) a nearest neighbor classifier based on the well-known and effective Dynamic Time Warping (DTW) distance measure. 
We also observe that a classifier trained on embeddings from Timenet give competitive results in comparison to a DTW-based classifier even when using significantly smaller set of labeled training data, providing further evidence that Timenet embeddings are robust. 
Finally, t-SNE visualizations of Timenet embeddings show that time series from different classes form well-separated clusters.
\end{abstract}

\section{Introduction}
Event detection from sensor data is of great importance to engineers who are concerned with heavy industrial equipments such as cranes, water pumps, and pulverizers, for making critical design, engineering and operational decisions. The current industry practice is to use unsupervised heuristic driven domain-based models for such tasks. Rampant use of diagnostic trouble codes in automobile industry,  Mahalanobis Taguchi method for anomaly detection in power-plants \citep{mtsbook}, and many other purpose specific heuristics such as a deceleration exceeding half of gravitational acceleration being termed as ``hard stop'' in vehicles \citep{hassan2015multi} are examples of such industry practices. Such approaches remain in mainstream use primarily because of unavailability of labeled time series data. However, with the advent of ``Industrial Internet'', unlabeled time series data from sensors is available in abundance. In this paper, we intend to explore the possibility of learning a model from the unlabeled time series data, which could be used to transform time series to \textit{representations} that are useful for further analysis, and call this model \textit{Timenet}. 

Learning good representations from data is an important task in machine learning \citep{bengio2013representation}. Recently, fixed-dimensional vector representations for words \citep{mikolov2013distributed} and for variable length sequences of words in the form of sentences, paragraphs, and documents
have been successfully used for natural language processing tasks such as machine translation and sentiment analysis \citep{le2014distributed,kiros2015skip}. Noticeably, deep Convolutional Neural Networks trained on millions of images from 1000 object classes from ImageNet \citep{russakovsky2015imagenet} have been used as off-the-shelf feature extractors to yield powerful generic image descriptors for a diverse range of tasks such as image classification, scene recognition and image retrieval \citep{sharif2014cnn, simonyan2014very}. These features or representations have even been shown to outperform models heavily tuned for the specific tasks.

Deep recurrent neural networks (RNNs) have been shown to perform hierarchical processing of time series with different layers tackling different time scales \citep{hermans2013training,LSTM_AD}. This serves as a motivation to explore a multilayered RNN as a Timenet model that processes and transforms a univariate time series to a fixed-dimensional vector representation, and once trained on diverse enough time series, serves as a generic off-the-shelf feature extractor for time series. To learn such a Timenet in an unsupervised manner, we leverage sequence-to-sequence (seq2seq) models \citep{p:seq2seq,p:seq2seqNIPS2014}: a seq2seq model consists of an encoder RNN and a decoder RNN. 
The encoder RNN takes a sequence as input and encodes it to a fixed-dimensional vector representation given by the final hidden state of the encoder, and the decoder RNN decodes this vector representation to generate another sequence as output. 
We train a seq2seq model on time series of varying length from diverse domains, and once trained, freeze the encoder RNN to be used as Timenet (refer Section \ref{sec:SAE} for details). 
Seq2seq models have been proposed for time series modeling tasks such as audio segment representation \citep{chung2016unsupervised}, anomaly detection from sensor data \citep{malhotra2016lstm}, and determining health of machines from sensor data \citep{malhotra2016multi}. 

To evaluate Timenet, we compare the embeddings obtained from Timenet with the embeddings given by the encoder RNN of a domain-specific SAE model on several time series classification (TSC) datasets from UCR Time Series Classification Archive \citep{UCRArchive}. For many datasets not used for training the Timenet, the embeddings from Timenet yield better classification performance compared to embeddings from the encoder of an SAE trained specifically on the respective datasets, and also outperforms a Dynamic Time Warping (DTW) based nearest neighbor classifier (DTW-C) which is a popular benchmark for TSC, and has been shown to be competitive across domains \citep{ding2008querying}. We also perform qualitative analysis of the embeddings from Timenet using t-SNE \citep{tsne}. 

The contributions of this paper can be summarised as follows:
\begin{enumerate}
 \item We show that it is possible to train \textit{Timenet: a deep recurrent neural network that serves as an off-the-shelf generic feature extractor for time series}. For several time series classification datasets which are not used for training Timenet, the classifiers learnt over embeddings given by Timenet perform better compared to i) classifiers over embeddings from data-specific RNNs learnt in same manner as the Timenet, and ii) DTW-C (Section \ref{ssec:TNvsSAE}).
 \vspace{-0.5em}
 \item Even when using \textit{significantly lesser amount of labeled training data}, Timenet embeddings based classifier gives competitive classification performance compared to DTW-C, suggesting that the embeddings are robust (Section \ref{ssec:TNvsSAE}).
 \vspace{-0.5em}
 \item \textit{We show that Timenet captures important characteristics of time series.} Timenet based embeddings of time series belonging to different classes form well-separated clusters when visualized using t-SNE (Section \ref{ssec:tSNE}). Similarly, Timenet based embeddings of time series from datasets belonging to different domains are also well-separated.
\end{enumerate}

The rest of the paper is organized as follows: We briefly introduce multilayered (deep) RNNs in Section \ref{sec:prelim}. Next, we explain our approach for learning Timenet using seq2seq models in Section \ref{sec:SAE}, followed by detailed experimental evaluation of Timenet and comparison with data-specific encoders in Section \ref{sec:exp}. We discuss the related work in Section \ref{sec:related}, and conclude in Section \ref{sec:conclusion} with observations and possible future directions.
\section{Preliminary: Multilayered RNN with Dropout}\label{sec:prelim}
We briefly introduce deep RNNs with Gated Recurrent Units \citep{p:seq2seq} in the hidden layers with dropout for regularization \citep{pham2014dropout,zaremba2014recurrent}. Dropout is applied to non-recurrent connections ensuring that the state of any hidden unit is not affected. This is important in RNNs to allow information flow across time-steps. 
For $l$th hidden layer of a multilayered RNN with $L$ layers, the hidden state $\mathbf{h}_{t}^{l}$ is obtained from the previous hidden state $\mathbf{h}_{t-1}^{l}$ and the hidden state $\mathbf{h}_{t}^{l-1}$ of the layer $l-1$. 
The hidden state transition from $t-1$ to $t$ for layer $l$ is given by function $f$:
\vspace{-0.5em}
\begin{equation}
f: \mathbf{h}_{t}^{l-1}, \mathbf{h}_{t-1}^{l} \rightarrow \mathbf{h}_{t}^{l}
\end{equation}
The function $f$ is implemented through the following transformations iteratively for $t=1$ to $T$:
\begin{equation}\label{eq:reset_gate}
 reset\, gate: \mathbf{r}_t^l = \sigma(\mathbf{W}_r^l.[\mathbf{D}(\mathbf{h}_{t}^{l-1}),\mathbf{h}_{t-1}^l])
\end{equation}
\vspace{-1.8em}
\begin{equation}
 update\, gate: \mathbf{u}_t^l = \sigma(\mathbf{W}_u^l.[\mathbf{D}(\mathbf{h}_{t}^{l-1}),\mathbf{h}_{t-1}^l])
\end{equation}
\vspace{-1.5em}
\begin{equation}\label{eq:proposed_state}
 proposed\, state: \mathbf{\tilde h}_t^l = \tanh(\mathbf{W}_p^l.[\mathbf{D}(\mathbf{h}_{t}^{l-1}),\mathbf{r}_t\odot \mathbf{h}_{t-1}^l])
\end{equation}
\vspace{-1.5em}
\begin{equation}\label{eq:hidden_state}
 hidden\, state: \mathbf{h}_t^l = (1-\mathbf{u}_t^l)\odot \mathbf{h}_{t-1}^l+\mathbf{u}_t^l\odot\mathbf{\tilde h}_t^l
\end{equation}
where $\odot$ is Hadamard product, $[\mathbf{a},\mathbf{b}]$ is concatenation of vectors $\mathbf{a}$ and $\mathbf{b}$, $\mathbf{D}(.)$ is dropout operator that randomly sets the dimensions of its argument to zero with probability equal to dropout rate, $\mathbf{h}_{t}^{0}$ is the input $z_t$ at time-step $t$. $\mathbf{W}_r$, $\mathbf{W}_u$, and $\mathbf{W}_p$ are weight matrices of appropriate dimensions s.t. $\mathbf{r}_t^l, \mathbf{u}_t^l, \mathbf{\tilde h}_t^l$, and $\mathbf{h}_t^l$ are vectors in $\mathbb{R}^{c^l}$, where $c^l$ is the number of units in layer $l$. 
The sigmoid ($\sigma$) and $tanh$ activation functions are applied element-wise. 

\section{Learning Timenet using Sequence Auto-encoder}\label{sec:SAE}
\begin{figure}[h]
 \centering
 \includegraphics[trim={0cm 9cm 0cm 3cm},clip=True,scale=0.28]{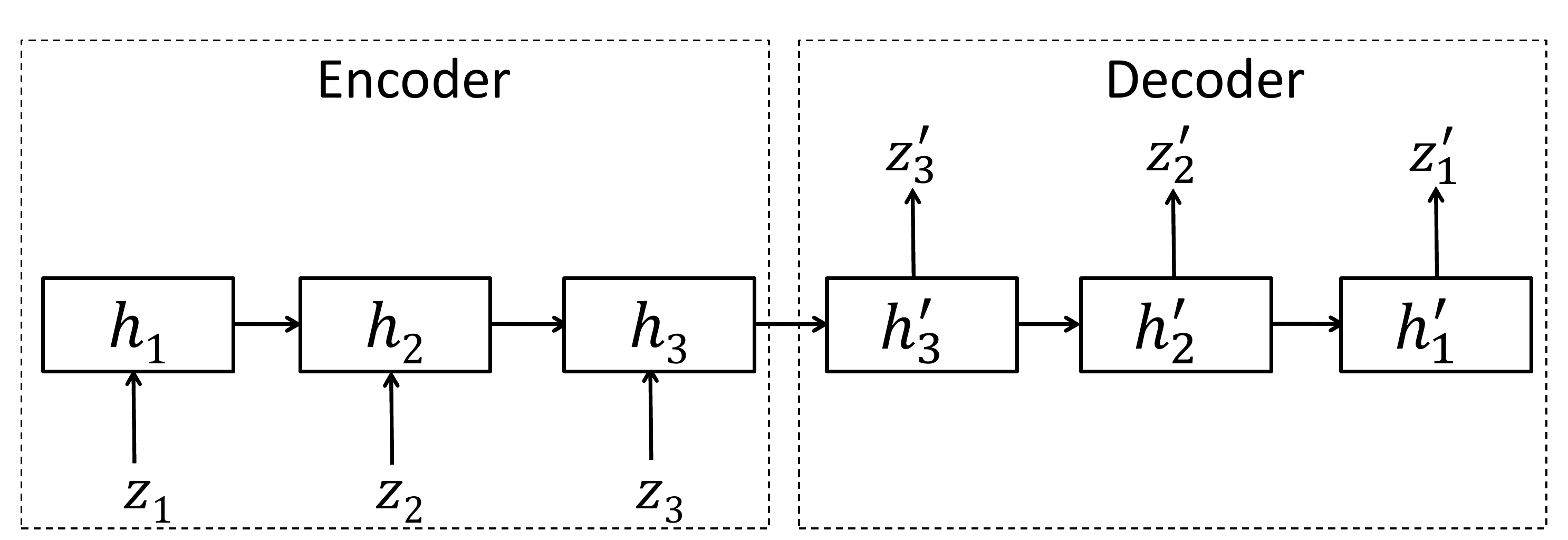}
 \caption{SAE for sample time series \{$z_1,z_2,z_3$\}. The final encoder state ($h_3$) is used as the embedding of the time series.}
 \label{fig:sae}
\end{figure}

We consider a sequence auto-encoder (SAE) based on seq2seq models (refer Figure \ref{fig:sae}). The SAE consists of two multilayered RNNs with Gated Recurrent Units in the hidden layers (as introduced in Section \ref{sec:prelim}): an encoder RNN and a decoder RNN. 
The encoder and decoder RNNs are trained jointly to minimize the reconstruction error on time series from diverse domains in an unsupervised manner. Once such an SAE is learnt, the encoder RNN from the encoder-decoder pair is used as a pre-trained Timenet model to obtain embeddings for test time series.

More specifically, given a time series $Z^{(i)} = \{z_1^{(i)}, z_2^{(i)}, ..., z_{T^{(i)}}^{(i)}\}$ of length $T^{(i)}$, $\mathbf{h}^{(i)}_{t}$ is the hidden state of the encoder at time $t$, where $\mathbf{h}^{(i)}_{t} \in \mathbb{R}^c$. The number of hidden units in the encoder is $c=\sum_{l=1}^{L}c^l$. 
The encoder captures information relevant to reconstruct the time series as it encodes the time series, and when it reaches the last point in the time series, the hidden state $\mathbf{h}^{(i)}_{T^{(i)}}$ is the vector representation or embedding for $Z^{(i)}$. 
The decoder has the same network structure as the encoder, with the final hidden state $\mathbf{h}^{(i)}_{T^{(i)}}$ of the encoder being used as the initial hidden state of decoder. The decoder additionally has a linear layer as the output layer. The decoder reconstructs the time series in reverse order, similar to \citep{p:seq2seqNIPS2014}, i.e., the target time series is $\{z_{T^{(i)}}^{(i)}, z_{T^{(i)}-1}^{(i)}, ..., z_{1}^{(i)}\}$. 
The SAE is trained to reconstruct the input time series so as to minimize the objective $E=\sum_{i=1}^{N}\sum_{t=1}^{T^{(i)}}(z_t^{(i)} - z'^{(i)}_t)^2$,
where $z'^{(i)}_t$ is the reconstructed value corresponding to $z^{(i)}_t$, $N$ is the number of time series instances. 

Unlike the conventional way of feeding an input sequence (\citep{p:seq2seq}) to the decoder during training as well as inference, the only inputs the decoder gets in our SAE model are the final hidden state of the encoder, and the steps $T$ for which the decoder has to be iterated in order to reconstruct the input. We observe that the embeddings or the final encoder states thus obtained carry all the relevant information necessary to represent a time series (refer Section \ref{sec:exp}). 

\textbf{Time complexity of obtaining embeddings from Timenet}: Consider a Timenet model with $L$ hidden layers, with each layer having $c$ recurrent units (without loss of generality). For any hidden layer $l>1$ and time-step $t$, the operations mentioned in Equations \ref{eq:reset_gate} - \ref{eq:proposed_state} involve multiplication of a weight matrix of dimensions $c \times 2c$ with a hidden state vector of size $2c \times 1$. 
For $l=1$, the weight matrix having dimension $c\times(c+1)$ is multiplied with vector of size $c+1$.
Therefore, the number of computations across layers is $O(Lc^2)$.
Equation \ref{eq:hidden_state} involves element-wise product of $c$-dimensional vectors, and hence involves $O(Lc)$ operations. 
Therefore, for a time series of length $T$, the total number of computations are $O(Lc^2T)$, which is linear in time series length $T$ for a given Timenet architecture.

\section{Experimental Evaluation}\label{sec:exp}
We begin with the details of how Timenet and data-specific SAEs were trained in Section \ref{ssec:details_timenet}; and then present a qualitative analysis of the ability of Timenet to learn robust embeddings using t-SNE in Section \ref{ssec:tSNE}. We compare Timenet embeddings with data-specific SAE embeddings on several TSC datasets in Section \ref{ssec:TNvsSAE}.
The time series datasets used are taken from UCR TSC Archive \citep{UCRArchive}, which is a source of large number of univariate time series datasets belonging to diverse domains. 
We also evaluate TimeNet on an industrial telematics dataset consisting of hourly readings from six sensors installed on engines in vehicles, where the task is to classify normal and abnormal behavior of engines (referred as Industrial Multivariate in Fig. \ref{fig:tSNE_mv_engine} and Table \ref{tab:results}). 

\subsection{Timenet and data-specific SAE training details}\label{ssec:details_timenet}
We chose 18 datasets for training, 6 datasets for validation, and another 30 as test datasets for Timenet evaluation from the UCR TSC Archive. The datasets are chosen such that time series length $T \leq 512$. Each dataset comes with a pre-defined train-test split and a class label for each time series. The training dataset is diverse as it contains time series belonging to 151 different classes from the 18 datasets with $T$ varying from 24 to 512 (refer Table \ref{tab:datasets} for details). When using a dataset as part of training or validation set, all the train and test time series from the dataset are included in the set. 
Each time series is normalized to have zero mean and unit variance.

We use Tensorflow \citep{abadi2015tensorflow} for implementing our models, and use Adam optimizer \citep{p:adamOpt} for training. We use learning rate 0.006, batch size 32, and dropout rate 0.4. All hidden layers have same number of hidden units ($c^l$). 
The best architecture is selected to have minimum average reconstruction error on the time series in the validation set. The best Timenet model obtained has $c^l=60$ and $L=3$ such that the embedding dimension $c=180$. This model takes about 17 hours to train for about 7k iterations on a NVIDIA Tesla K40C 12GB GPU. Data-specific SAE models are trained under same parameter settings while tuning for $c^l$ and $L$, in an unsupervised manner, using only the training set of the respective dataset.

\subsection{Visualization of embeddings using t-SNE}\label{ssec:tSNE}
\begin{figure}[h]
    \centering
    \begin{subfigure}[b]{0.23\textwidth}
        \includegraphics[trim={2.5cm 2.5cm 2.5cm 2.5cm},clip=True,width=\textwidth]{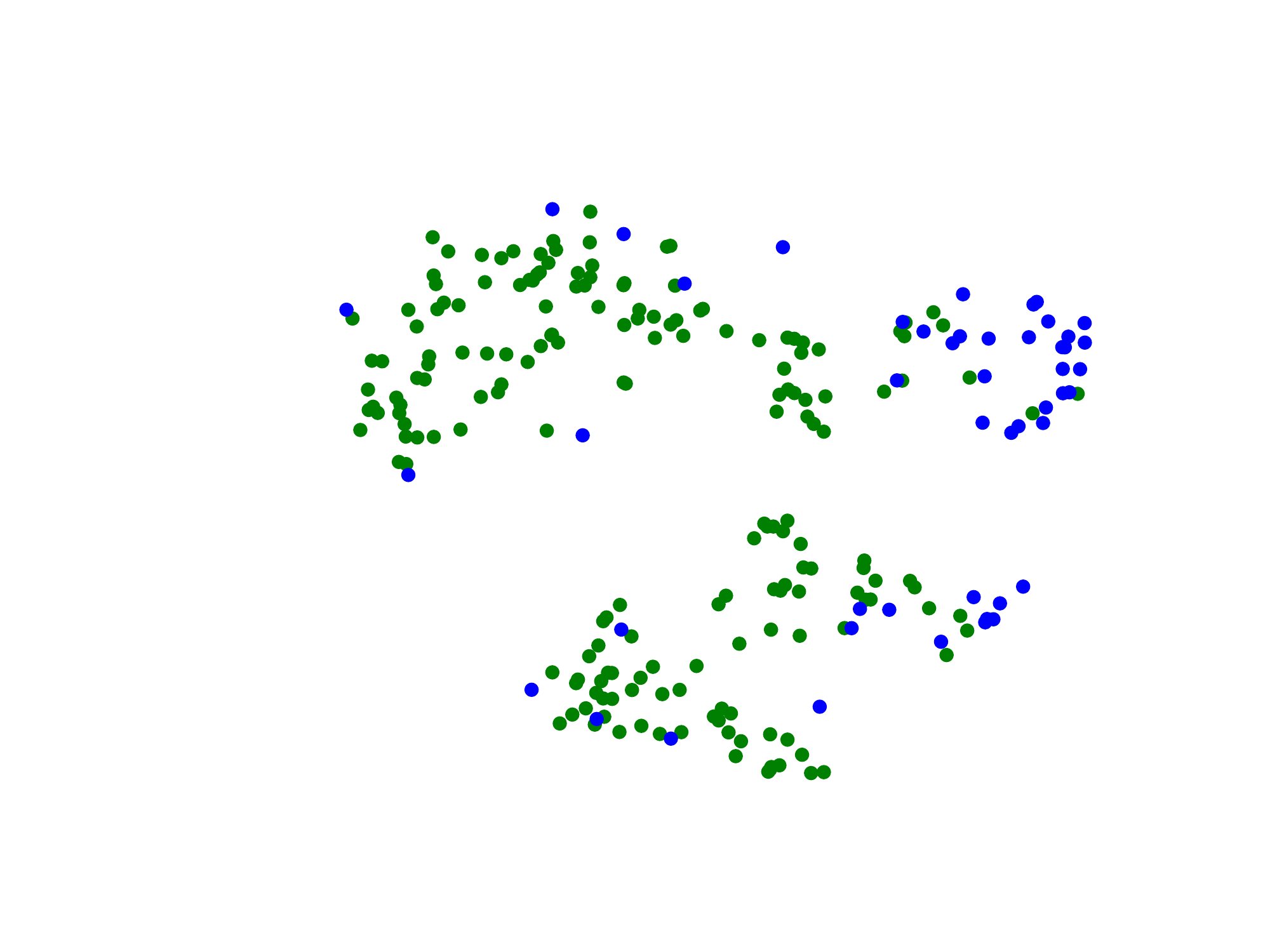}
        \subcaption{Industrial Multivariate (C=2)}
        \label{fig:tSNE_mv_engine}
    \end{subfigure}
    \begin{subfigure}[b]{0.23\textwidth} 
        \includegraphics[trim={2.5cm 2.5cm 2.5cm 2.5cm},clip=True,width=\textwidth]{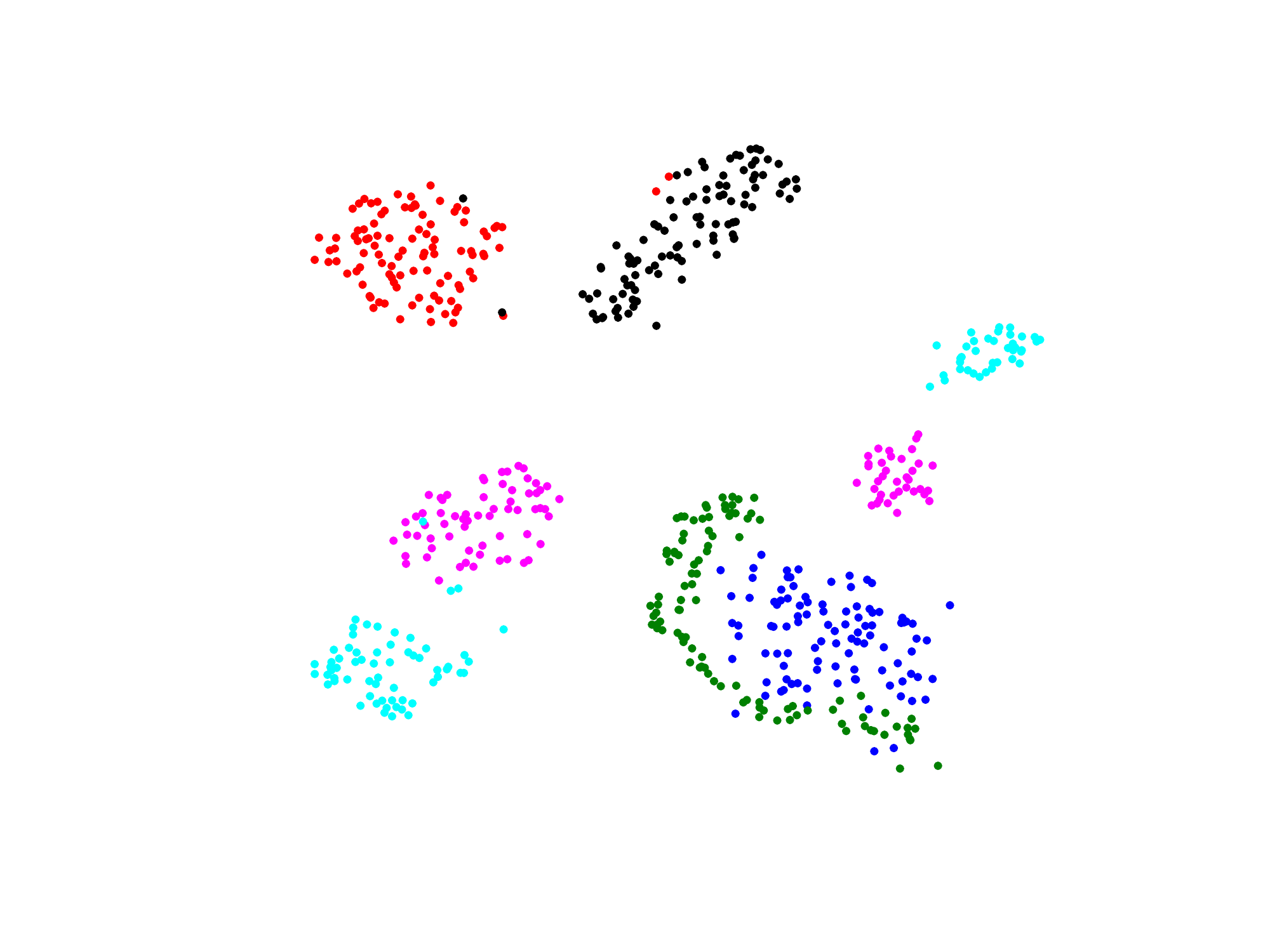}
	\subcaption{Synthetic Control (C=6)}
	\label{fig:tSNE_synt_cont}
    \end{subfigure}
    \begin{subfigure}[b]{0.23\textwidth}
        \includegraphics[trim={2.5cm 2.5cm 2.5cm 2.5cm},clip=True,width=\textwidth]{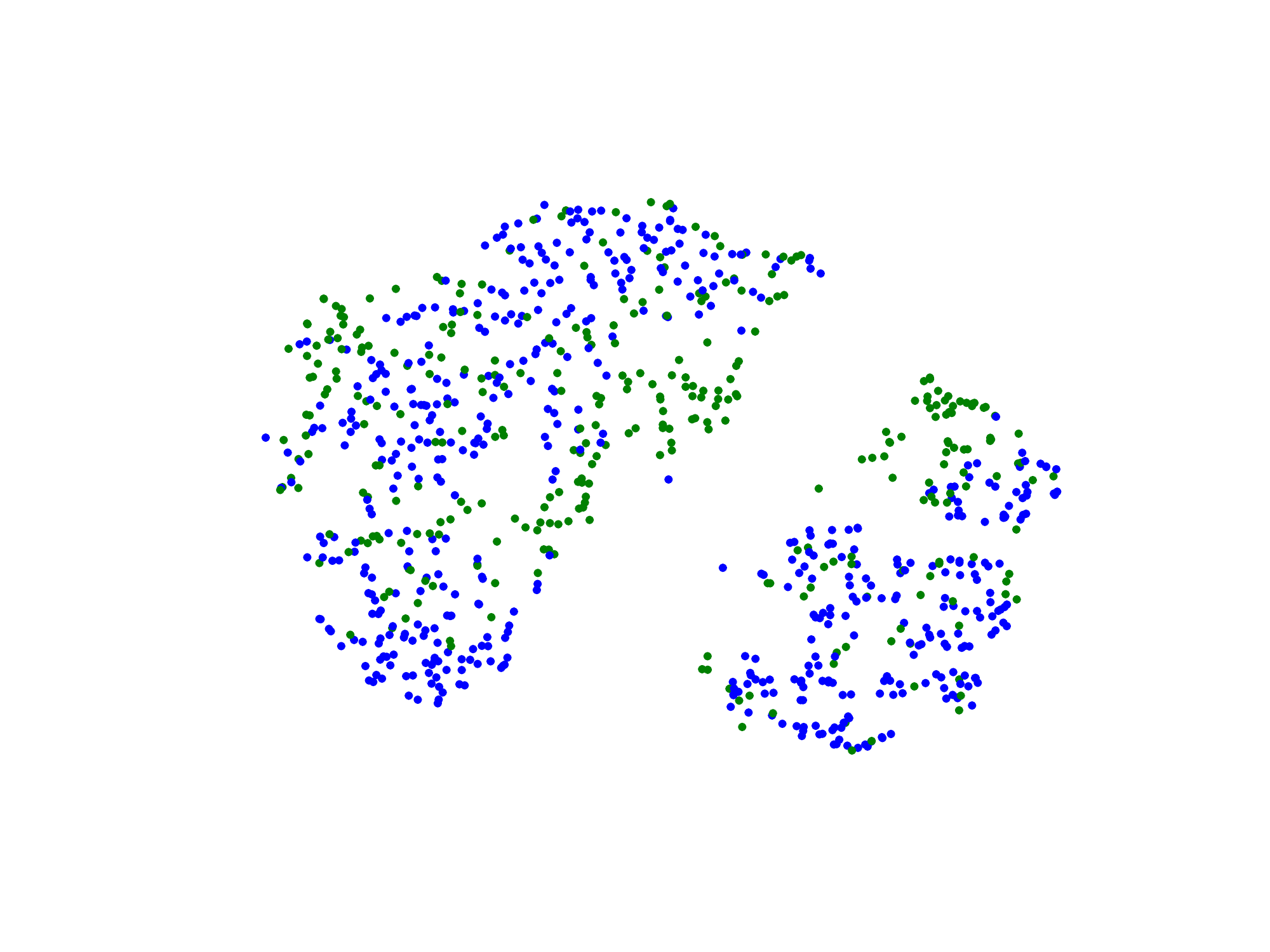}
        \subcaption{MiddlePhalanxOutlineCorrect (C=2)}
    \end{subfigure}
    \begin{subfigure}[b]{0.23\textwidth}
        \includegraphics[trim={2.5cm 2.5cm 2.5cm 2.5cm},clip=True,width=\textwidth]{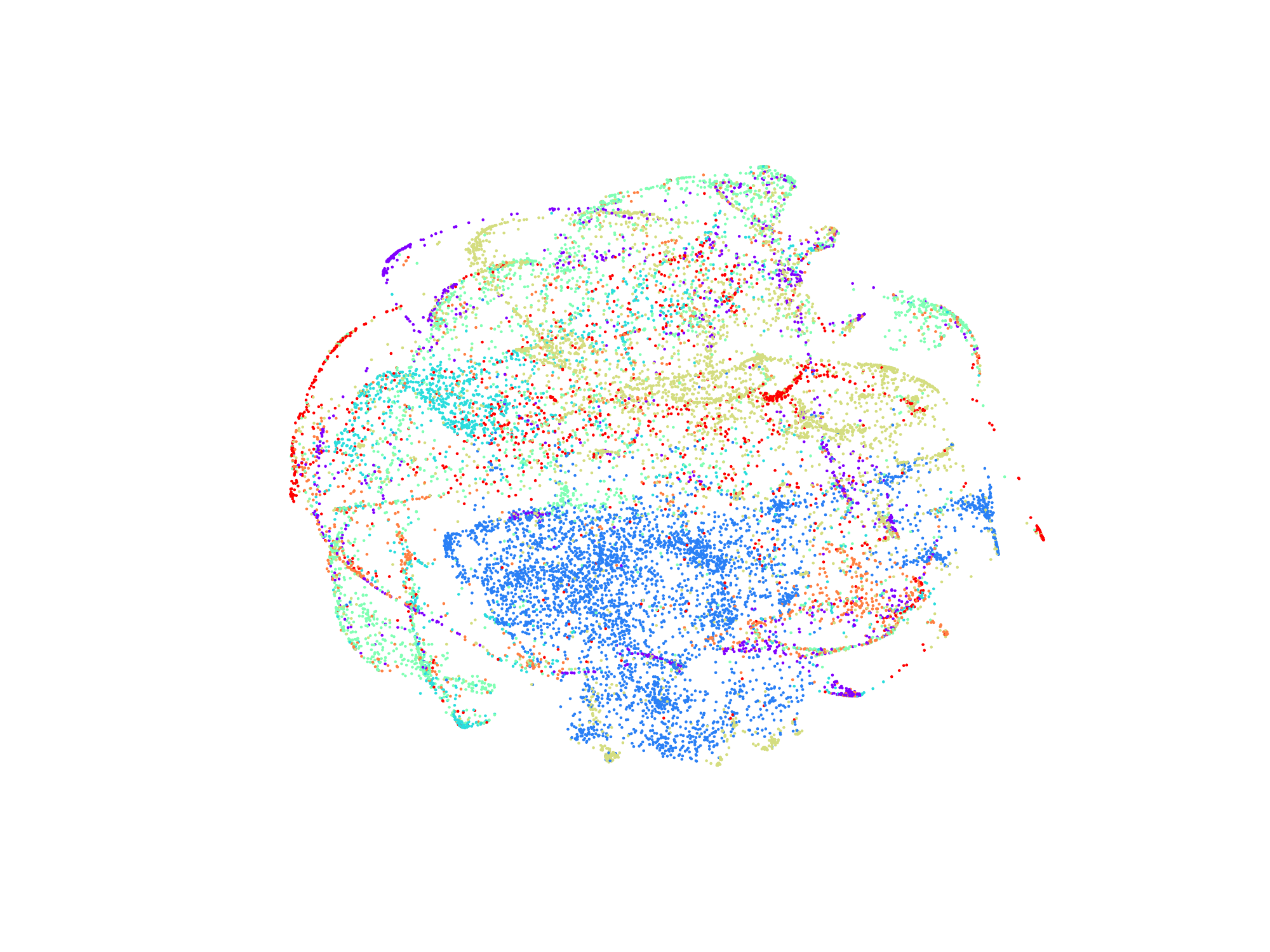}
        \subcaption{Electric Devices (C=7)}
    \end{subfigure}
    \begin{subfigure}[b]{0.23\textwidth}
        \includegraphics[trim={2.5cm 2.5cm 2.5cm 2.5cm},clip=True,width=\textwidth]{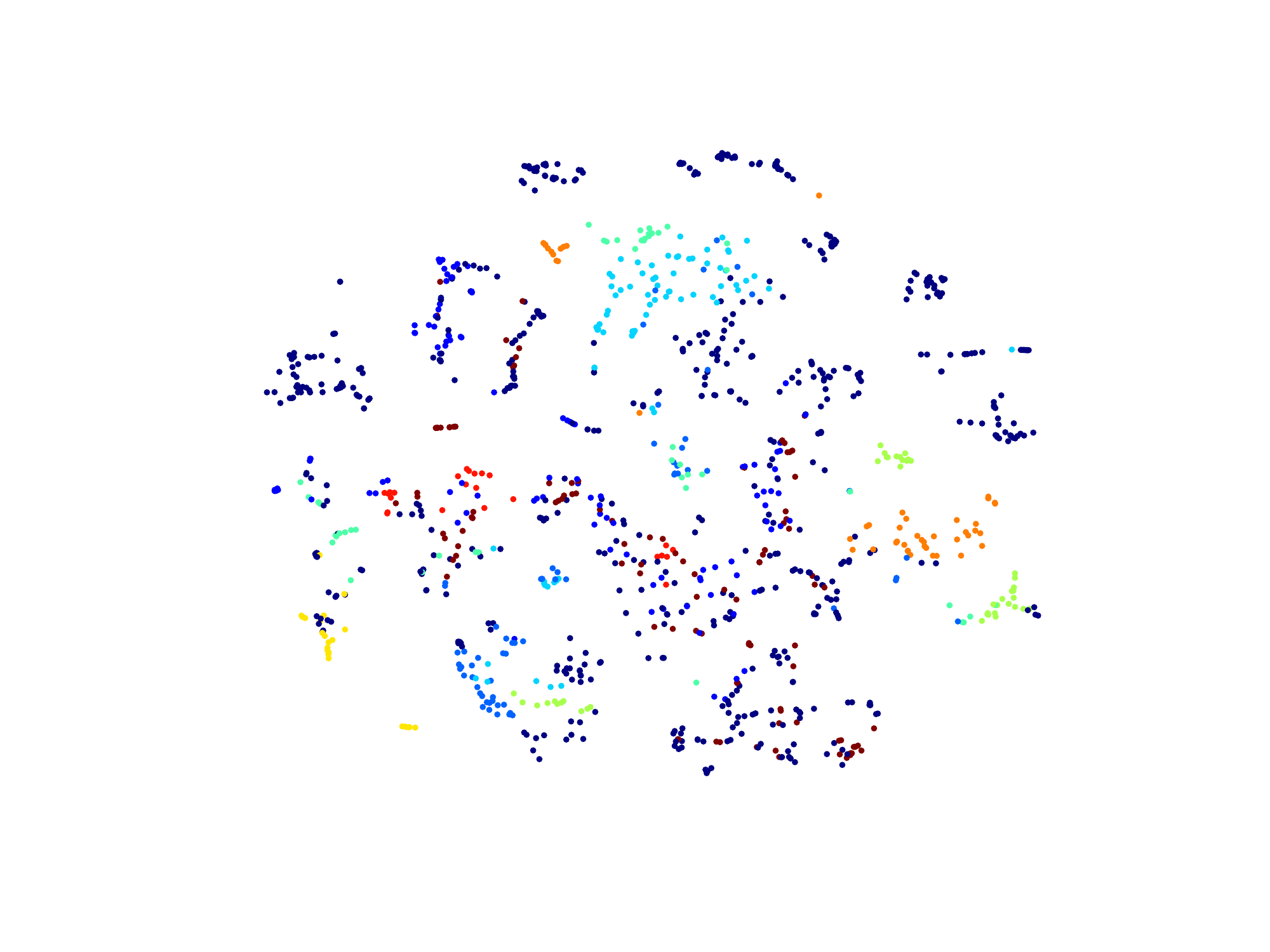}
	\subcaption{MedicalImages (C=10)}
    \end{subfigure}
    \begin{subfigure}[b]{0.23\textwidth}
        \includegraphics[trim={2.5cm 2.5cm 2.5cm 2.5cm},clip=True,width=\textwidth]{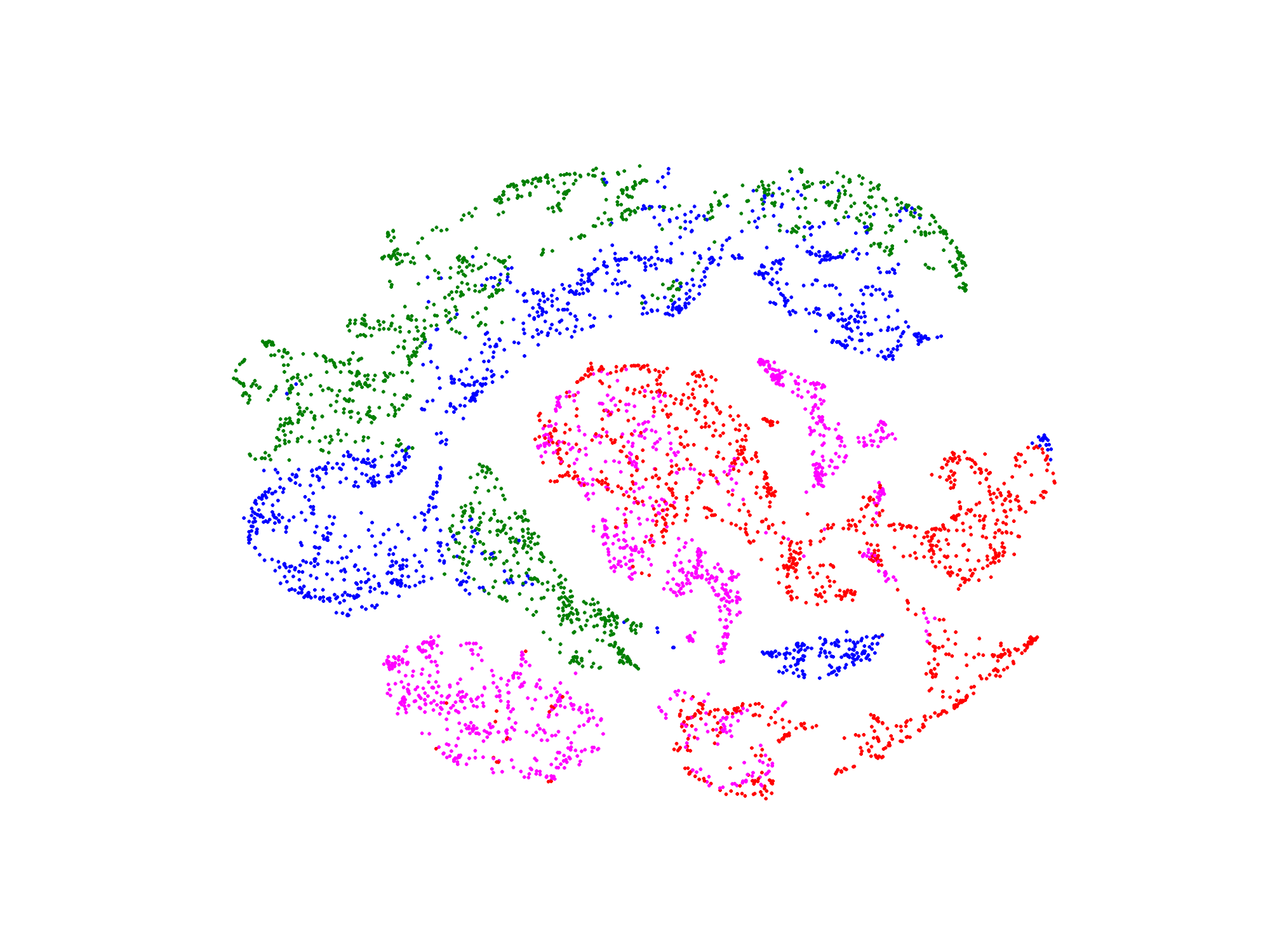}
        \subcaption{Two Patterns (C=4)}
    \end{subfigure}    
    \begin{subfigure}[b]{0.23\textwidth}
        \includegraphics[trim={2.5cm 2.5cm 2.5cm 2.5cm},clip=True,width=\textwidth]{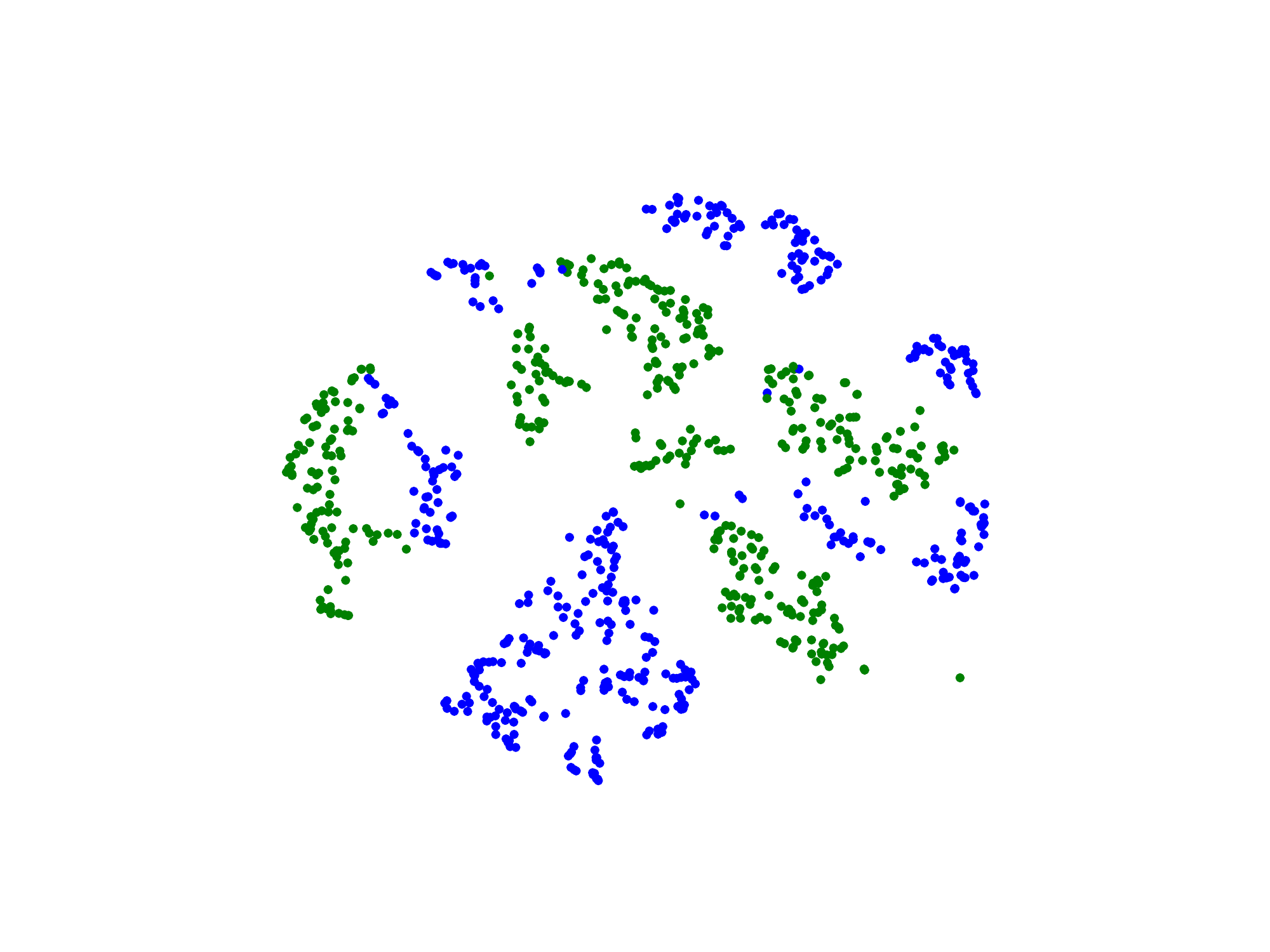}
        \subcaption{ECGFiveDays (C=2)}
    \end{subfigure}
    \begin{subfigure}[b]{0.23\textwidth}
        \includegraphics[trim={2.5cm 2.5cm 2.5cm 2.5cm},clip=True,width=\textwidth]{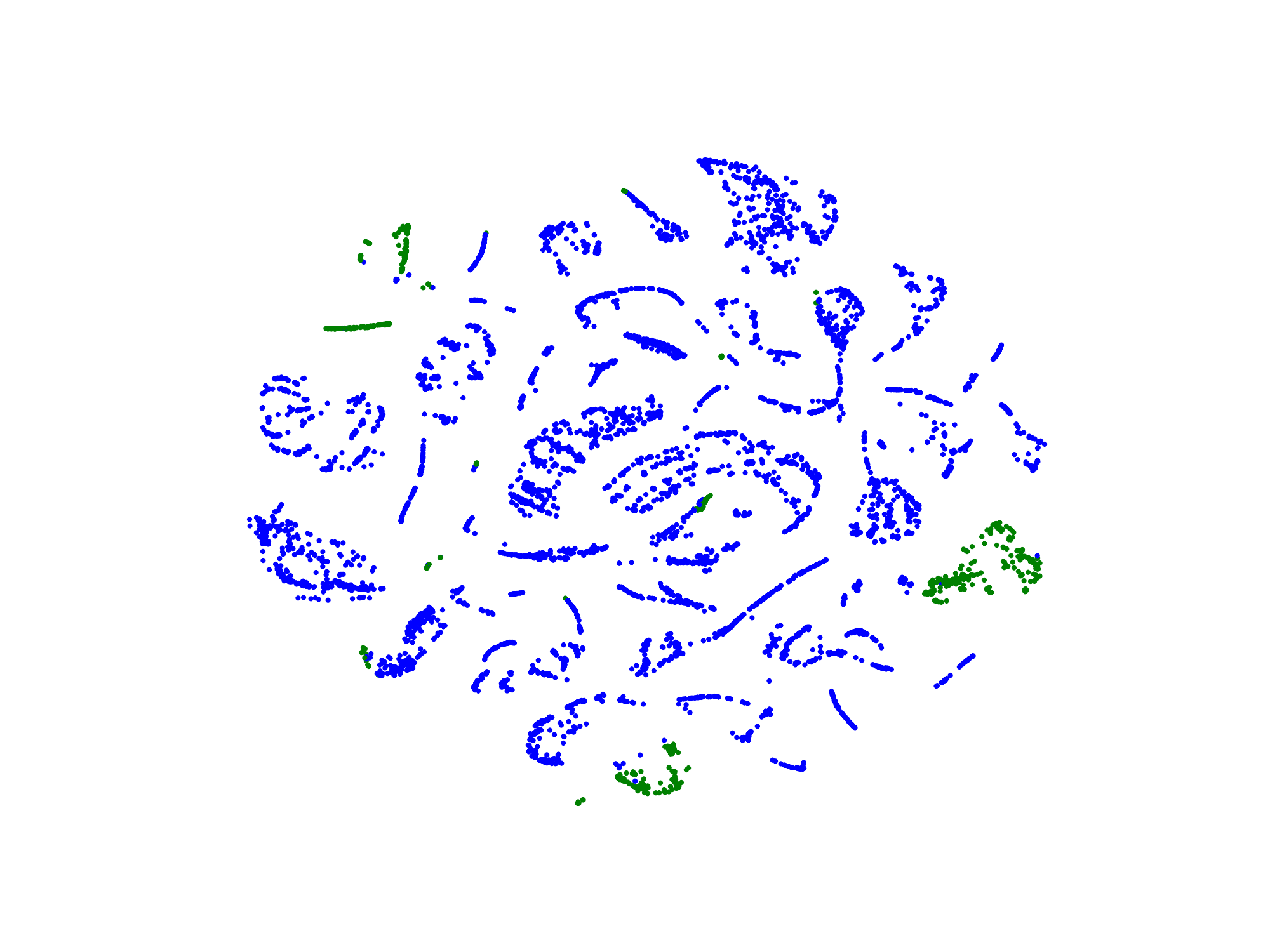}
	\subcaption{Wafer (C=2)}
    \end{subfigure}
    \begin{subfigure}[b]{0.23\textwidth}
        \includegraphics[trim={2.5cm 2.0cm 2.5cm 2.5cm},clip=True,width=\textwidth]{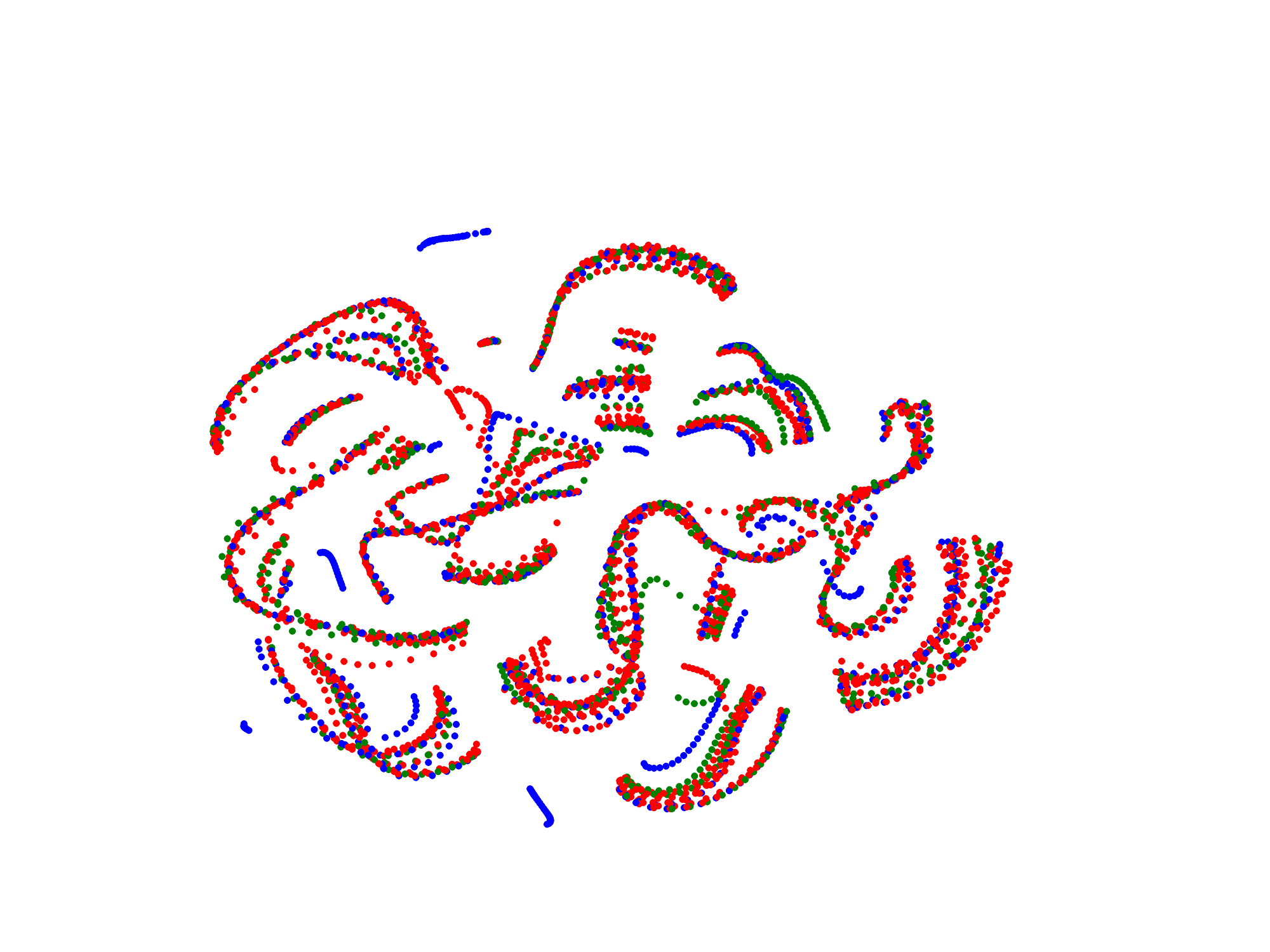}
	\subcaption{ChlorineConcentration (C=3)}
    \end{subfigure}    
    \begin{subfigure}[b]{0.23\textwidth}
        \includegraphics[trim={2.0cm 2.0cm 2.0cm 2.0cm},clip=True,width=\textwidth]{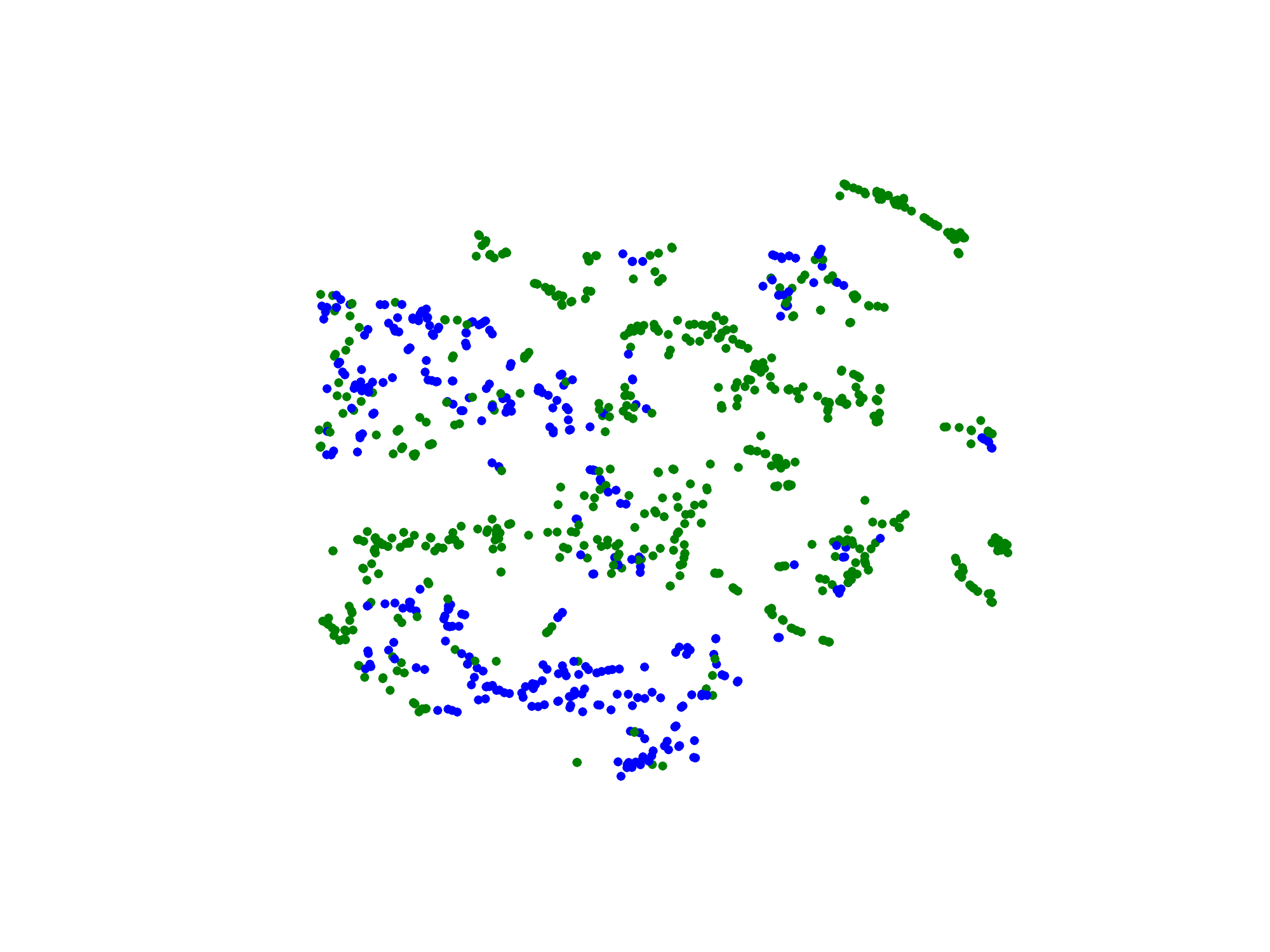}
        \subcaption{Strawberry (C=2)}
    \end{subfigure}
    \begin{subfigure}[b]{0.23\textwidth}
        \includegraphics[trim={2.5cm 2.5cm 2.5cm 2.5cm},clip=True,width=\textwidth]{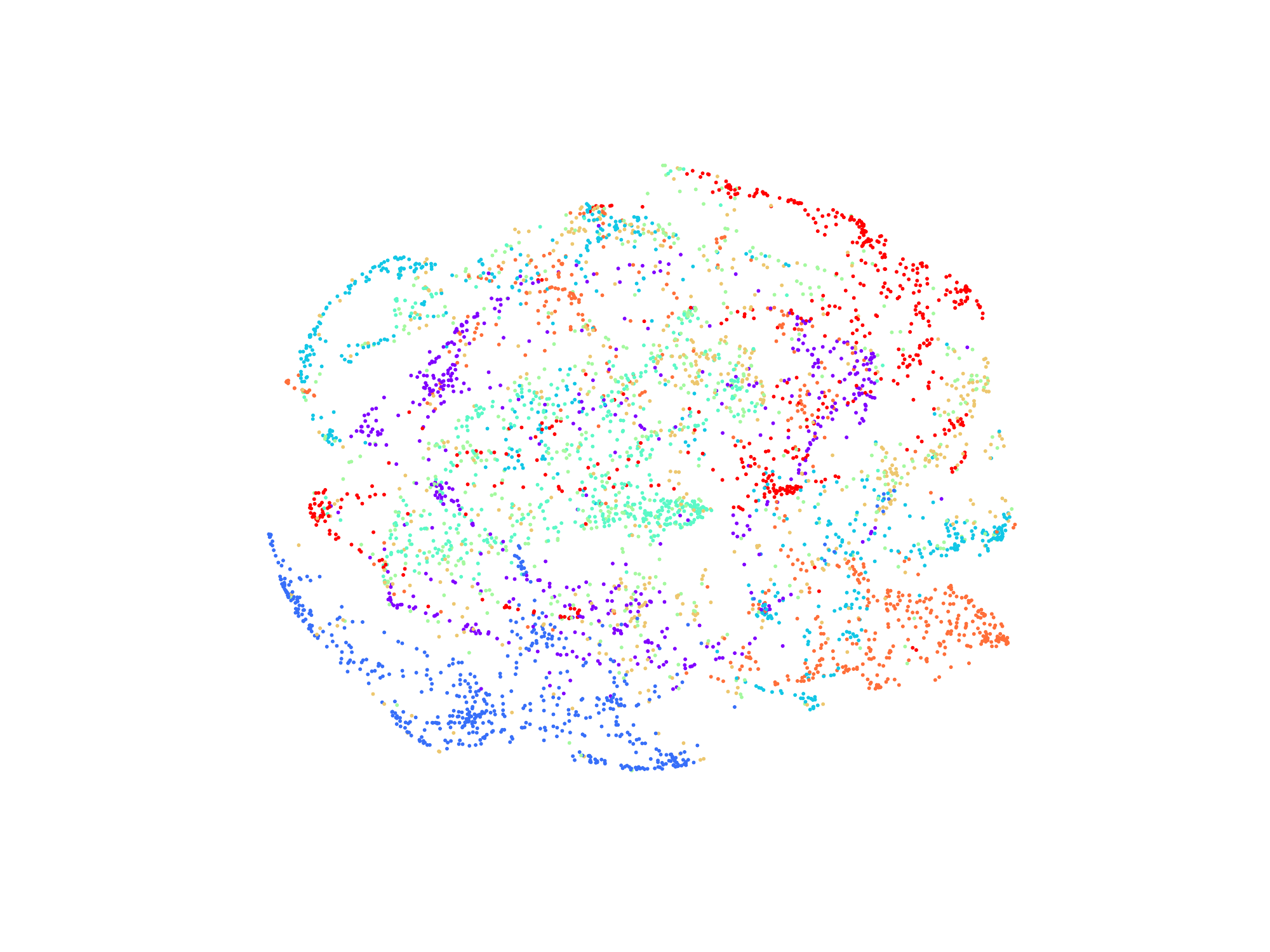}
        \subcaption{uWaveGestureLibrary\_X (C=8)}
    \end{subfigure}
    \begin{subfigure}[b]{0.23\textwidth}
        \includegraphics[trim={2.5cm 2.5cm 2.5cm 2.5cm},clip=True,width=\textwidth]{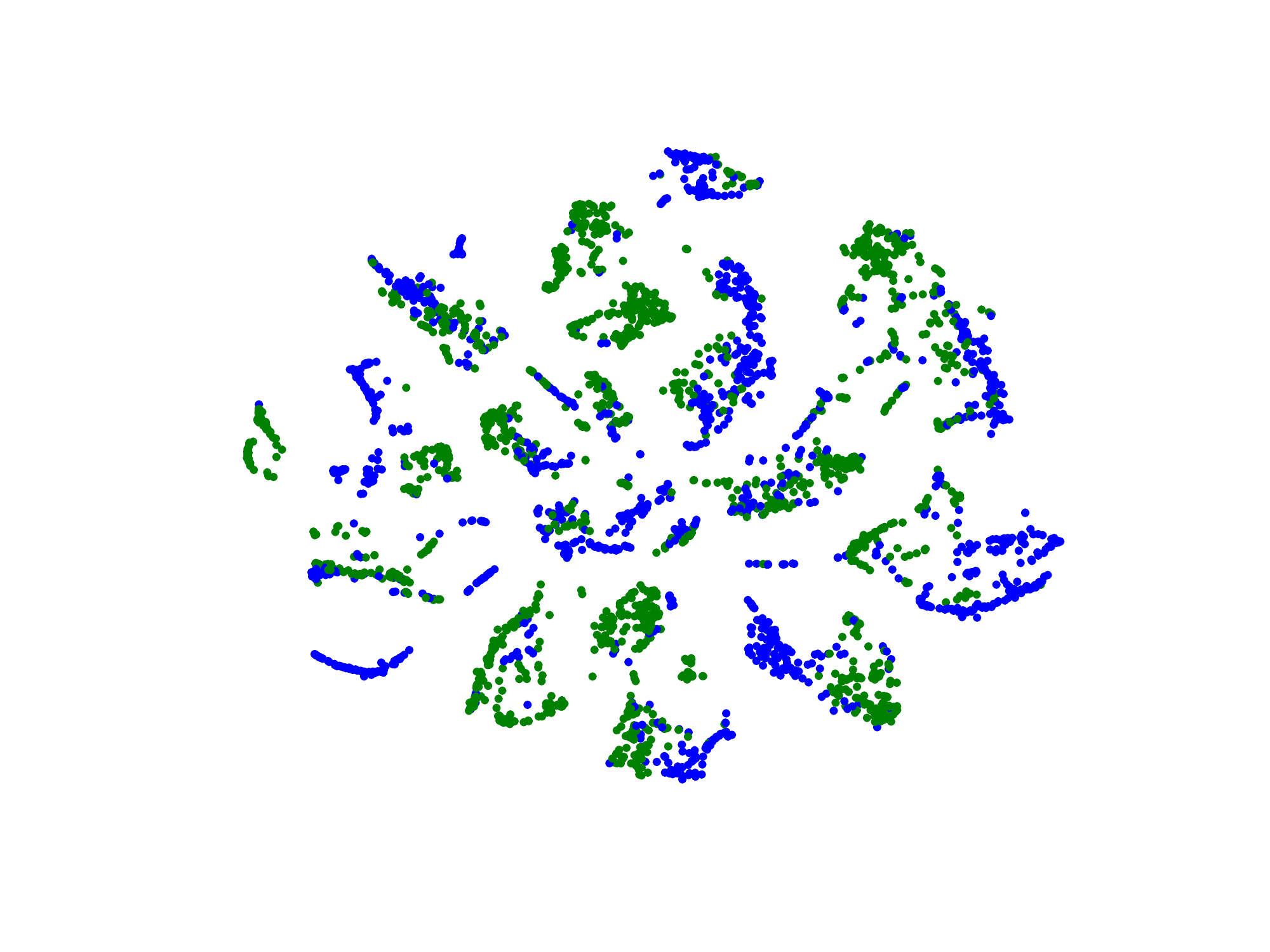}
        \subcaption{Yoga (C=2)}
        \label{fig:tSNE_yoga}
    \end{subfigure}
    \begin{subfigure}[b]{0.23\textwidth}
        \includegraphics[trim={2.5cm 2.5cm 2.5cm 2.5cm},clip=True,width=\textwidth]{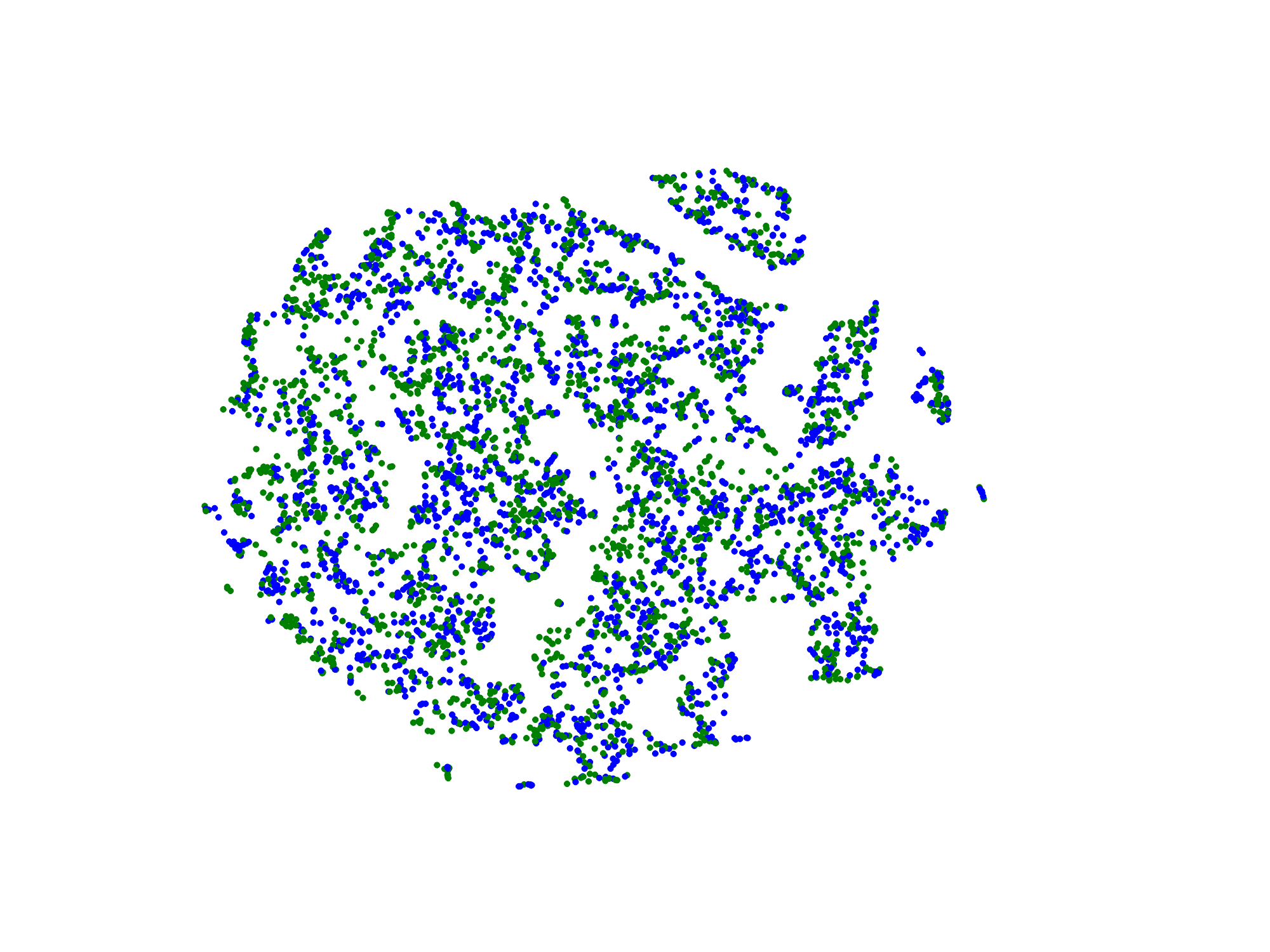}
	\subcaption{FordA (C=2)}
	\label{fig:FordA}
    \end{subfigure}
    \begin{subfigure}[b]{0.23\textwidth}
        \includegraphics[trim={2.5cm 2.5cm 2.5cm 2.5cm},clip=True,width=\textwidth]{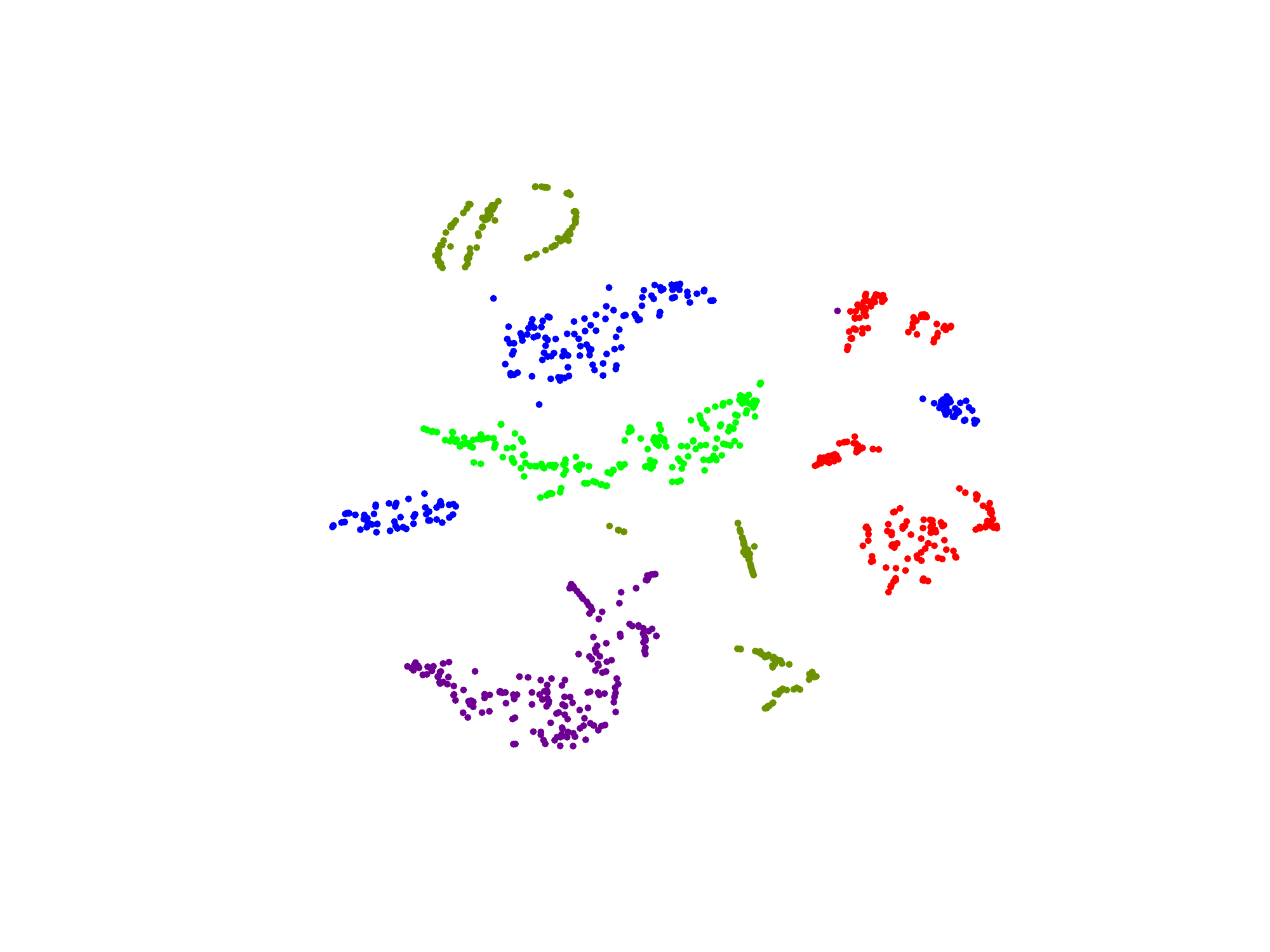}
        \subcaption{Five datasets - i.}
        \label{fig:tSNE_alltest1}
    \end{subfigure}
    \begin{subfigure}[b]{0.23\textwidth}
        \includegraphics[trim={2.5cm 2.5cm 2.5cm 2.5cm},clip=True,width=\textwidth]{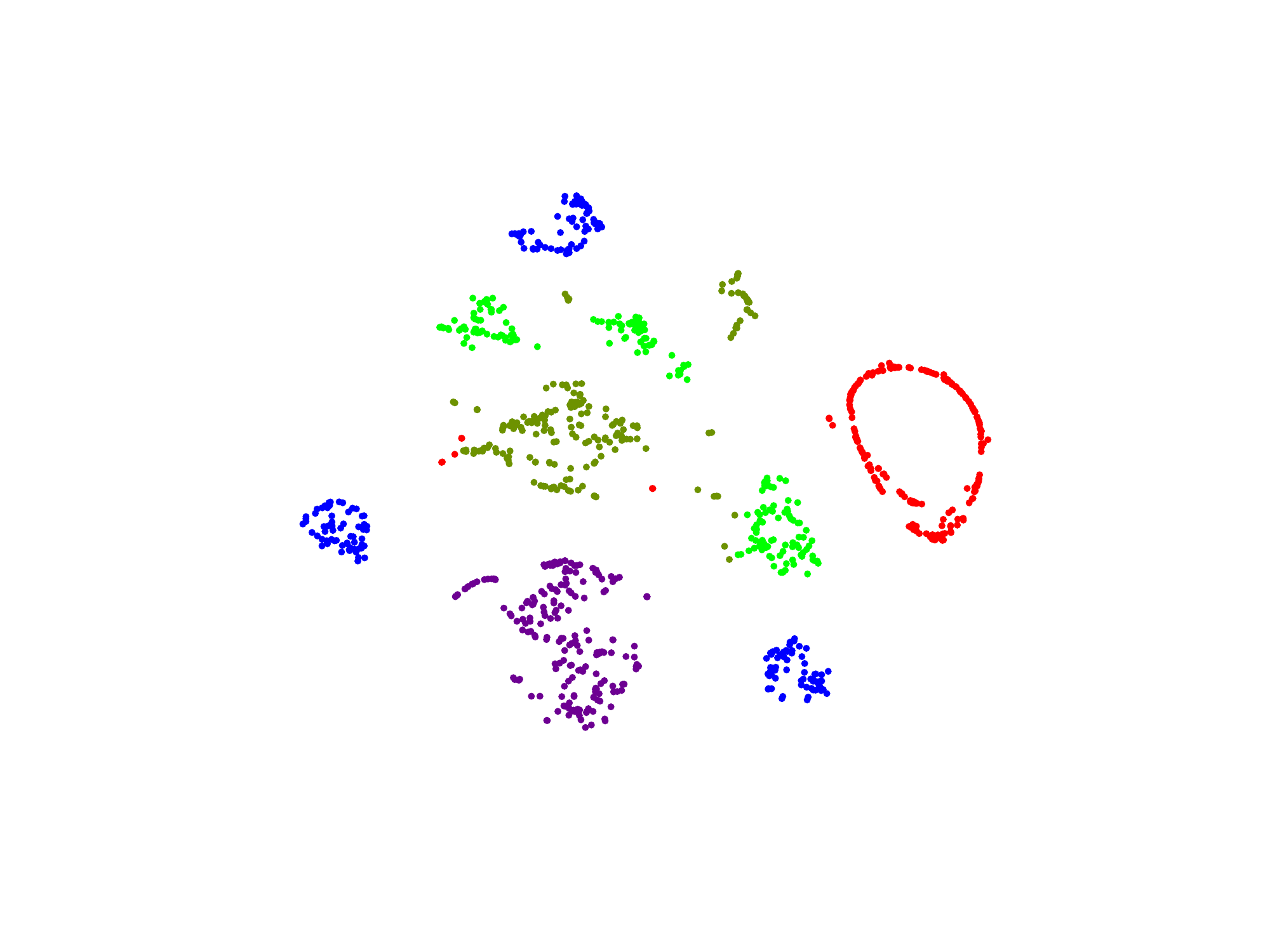}
        \subcaption{Five datasets - ii.}
    \end{subfigure}
    \begin{subfigure}[b]{0.23\textwidth}
        \includegraphics[trim={2.5cm 2.5cm 2.5cm 2.5cm},clip=True,width=\textwidth]{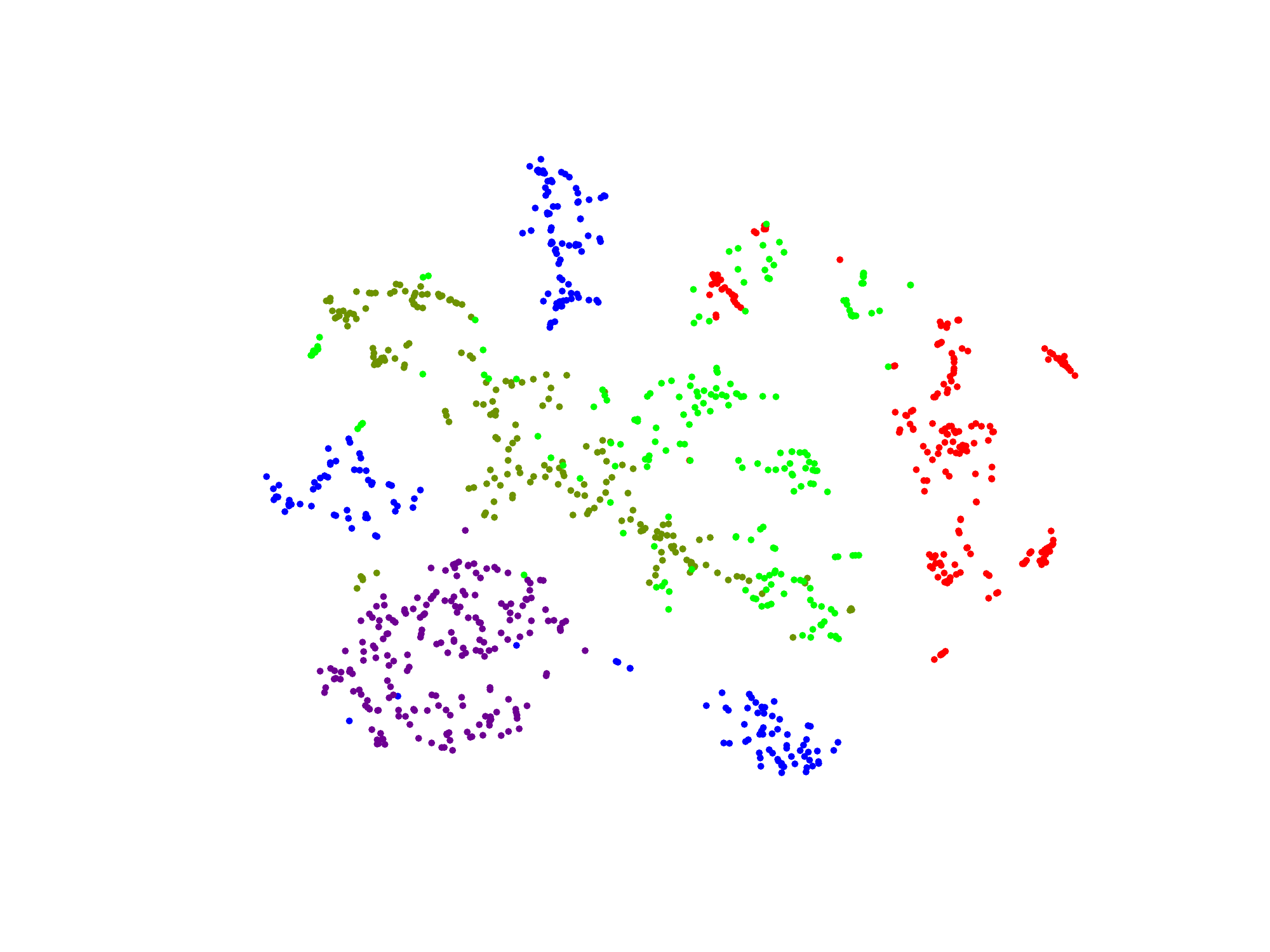}
        \subcaption{Five datasets - iii.}
    \end{subfigure}
    \begin{subfigure}[b]{0.23\textwidth}
        \includegraphics[trim={2.5cm 2.5cm 2.5cm 2.5cm},clip=True,width=\textwidth]{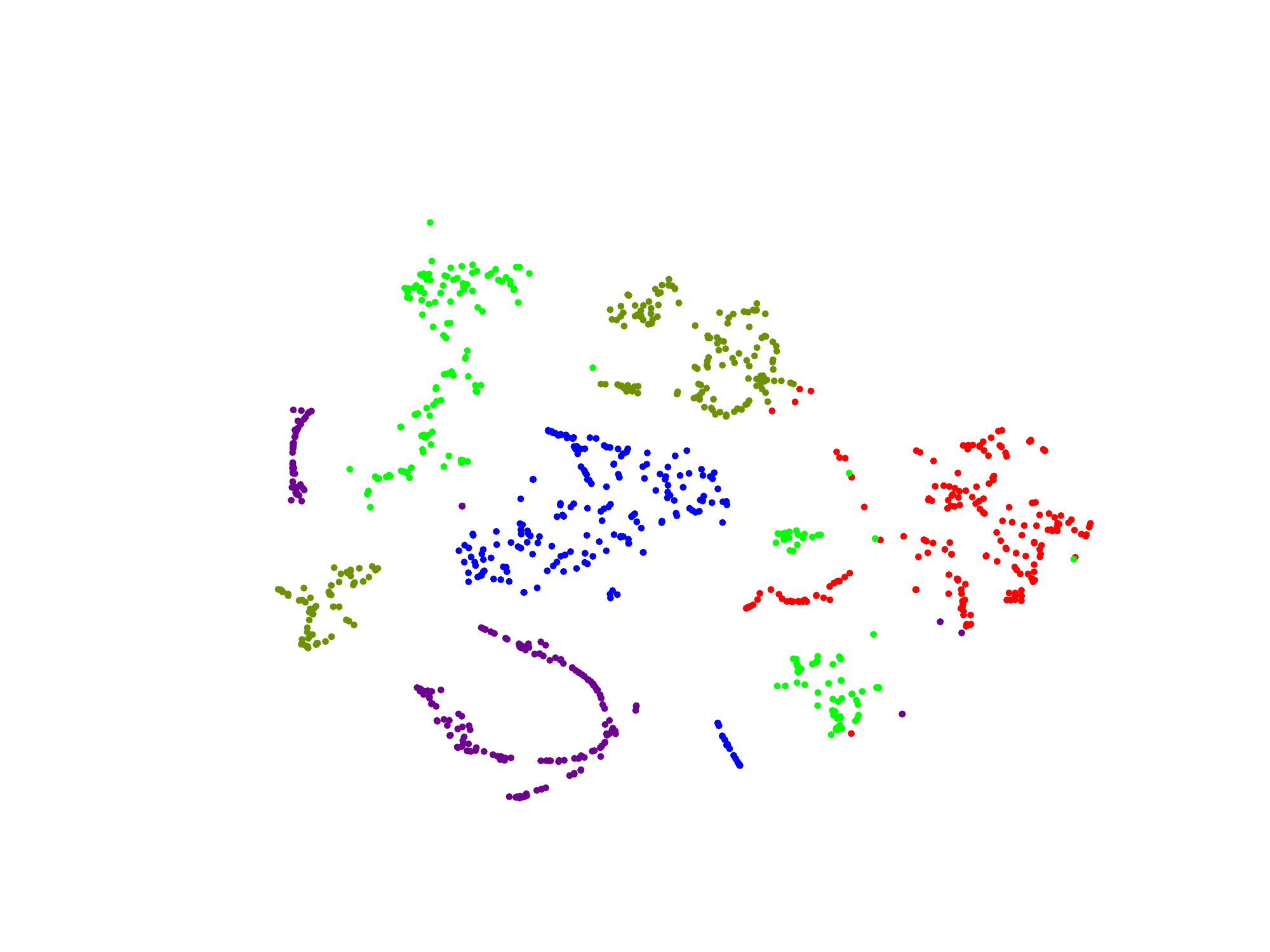}
        \subcaption{Five datasets - iv.}
        \label{fig:tSNE_alltest4}
    \end{subfigure}
    \caption{Timenet embeddings visualized using t-SNE. Image best viewed magnified.}
\end{figure}

Figures \ref{fig:tSNE_mv_engine} - \ref{fig:FordA} show t-SNE visualizations of embeddings obtained from Timenet for test datasets (refer Table \ref{tab:results}) with number of classes $C\leq10$ (for visual clarity). For each test dataset, the embeddings for all the time series from both train and test splits are shown. Due to space constraints, we show t-SNE visualization for one dataset each (selected randomly) from the ten Phalanx datasets and two ECG datasets (as listed in Table \ref{tab:results}). To compare the embeddings obtained from Timenet across test datasets, we consider 200 randomly selected time series each from five randomly selected test datasets, and visualize them using t-SNE. Figures \ref{fig:tSNE_alltest1} - \ref{fig:tSNE_alltest4} show t-SNE plots for four such random selections. We  observe that embeddings for time series belonging to different classes within a dataset form well-separated clusters in most cases (Figures \ref{fig:tSNE_mv_engine} - \ref{fig:FordA}). Further, the embeddings for time series from one dataset are well-separated from embeddings for time series from other datasets (Figures \ref{fig:tSNE_alltest1} - \ref{fig:tSNE_alltest4}). This suggests that \textit{Timenet is able to capture important characteristics of time series in the embeddings produced.}

\subsection{Embeddings from Timenet and data-specific SAE}\label{ssec:TNvsSAE}
For each test dataset, we learn two non-linear Support Vector Machine (SVM) \citep{burges1998tutorial} classifiers with radial basis function (RBF) kernel: i) TN-C: using Timenet embeddings as features, ii) SAE-C: using embeddings obtained from the encoder of a data-specific SAE model as features. 
It is to be noted that none of the 30 test datasets on which we evaluate Timenet are used for training the Timenet. However, for training a data-specific SAE we use the training set of the respective dataset (without using the class label information).
The parameters C (cost) and $\gamma$ of the RBF kernel of SVM are selected using 5-fold cross-validation over the logarithmic grid of $10^{-3}$ to $10^3$.
The classification error rates for TN-C, SAE-C, and DTW-C are reported in Table \ref{tab:results}. The error rates for DTW-C are taken from \citep{lines2015time}.
\begin{table}[h]
  \scriptsize
  \centering
  \begin{tabular}{|p{5.25cm}|c|p{0.77cm}|p{0.7cm}|p{0.7cm}|}\hline
    \textbf{Dataset} & \textbf{T} & \textbf{DTW-C} & \textbf{SAE-C} & \textbf{TN-C}\\ \hline
    \textbf{Industrial Multivariate} & 30 & - & 0.221 & \textbf{0.173} \\\hline 
    \textbf{Synthetic Control} & 60 & 0.017 & 0.017 & \textbf{0.013} \\\hline
    \textbf{PhalangesOutlinesCorrect} & 80 & 0.239 & 0.228  & \textbf{0.207}\\
    \textbf{DistalPhalanxOutlineAgeGroup*} & 80 & 0.228 & \textbf{0.160} & 0.223\\
    \textbf{DistalPhalanxOutlineCorrect*} & 80 & 0.232 & \textbf{0.187} & 0.188\\
    \textbf{DistalPhalanxTW*} & 80 & 0.272 & 0.243 & \textbf{0.208} \\
    \textbf{MiddlePhalanxOutlineAgeGroup*} & 80 & 0.253 & 0.348 & \textbf{0.210} \\
    \textbf{MiddlePhalanxOutlineCorrect*} & 80 & 0.318 & 0.307 & \textbf{0.270} \\
    \textbf{MiddlePhalanxTW*} & 80 & 0.419 & 0.381 & \textbf{0.363} \\
    \textbf{ProximalPhalanxOutlineAgeGroup*} & 80 & 0.215 & \textbf{0.137} & 0.146 \\
    \textbf{ProximalPhalanxOutlineCorrect*} & 80 & 0.210 & 0.179 & \textbf{0.175} \\
    \textbf{ProximalPhalanxTW*} & 80 & 0.263 & \textbf{0.188} & 0.195 \\ \hline
    \textbf{ElectricDevices} & 96 & 0.376 & 0.335 & \textbf{0.267} \\ \hline
    \textbf{MedicalImages} & 99 & 0.253 & \textbf{0.247} & 0.250\\ \hline
    \textbf{Swedish Leaf} & 128 & 0.157 & \textbf{0.099} & 0.102 \\ \hline
    \textbf{Two Patterns} & 128 & 0.002 & 0.001 & \textbf{0.000} \\ \hline
    \textbf{ECG5000} & 140 & 0.075 & \textbf{0.066} & 0.069 \\
    \textbf{ECGFiveDays*} & 136 & 0.203 & \textbf{0.063} & 0.074 \\ \hline
    \textbf{Wafer} & 152 & \textbf{0.005} & 0.006 & \textbf{0.005} \\ \hline
    \textbf{ChlorineConcentration} & 166 & 0.35 & 0.277  & \textbf{0.269} \\ \hline
    \textbf{Adiac} & 176 & 0.391 & 0.435 & \textbf{0.322} \\ \hline
    \textbf{Strawberry} & 235 & \textbf{0.062} & 0.070 & \textbf{0.062} \\ \hline
    \textbf{Cricket\_X} & 300 & \textbf{0.236} & 0.341 & 0.300 \\
    \textbf{Cricket\_Y*} & 300 & \textbf{0.197} & 0.397 & 0.338 \\
    \textbf{Cricket\_Z*} & 300 & \textbf{0.180} & 0.305 & 0.308 \\ \hline
    \textbf{uWaveGestureLibrary\_X} & 315 & 0.227 & \textbf{0.211} & 0.214 \\ 
    \textbf{uWaveGestureLibrary\_Y*} & 315 & 0.301 & \textbf{0.291} &0.311\\ 
    \textbf{uWaveGestureLibrary\_Z*} & 315 & 0.322 & \textbf{0.280} & 0.281 \\ \hline
    \textbf{Yoga} & 426 & \textbf{0.155} & 0.174 & 0.160 \\ \hline
    \textbf{FordA} & 500 & 0.341 & 0.284 & \textbf{0.219} \\
    \textbf{FordB} & 500 & 0.414 & 0.405 & \textbf{0.263} \\\hline \hline
    \textbf{Win or Tie compared to DTW-C} & &-& 22/30 & \textbf{25/30} \\ \hline    
  \end{tabular}
  \begin{tabular}{|p{0.7cm}|p{0.7cm}|p{0.7cm}|p{0.7cm}|}\hline
    \textbf{TN-C$_{2/3}$}&\textbf{TN-C$_{L1}$}&\textbf{TN-C$_{L2}$}&\textbf{TN-C$_{L3}$}\\ \hline
     0.176&0.135&0.154&0.154\\ \hline 
     0.016&0.010&0.013&0.027\\ \hline
     0.225&0.213&0.221&0.217\\
     0.211&0.178&0.200&0.165\\
     0.201&0.188&0.178&0.185\\
     0.220&0.203&0.213&0.223\\
     0.229&0.215&0.280&0.205\\
     0.344&0.475&0.472&0.295\\
     0.392&0.361&0.371&0.366\\
     0.154&0.141&0.151&0.156\\
     0.199&0.175&0.175&0.175\\
     0.194&0.200&0.200&0.188\\ \hline
     0.288&0.265&0.280&0.309\\ \hline
     0.271&0.238&0.246&0.232\\ \hline
     0.139&0.123&0.126&0.115\\ \hline
     0.002&0.000&0.002&0.007\\ \hline
     0.069&0.063&0.069&0.066\\
     0.150&0.129&0.127&0.096\\ \hline
     0.007&0.008&0.006&0.009\\ \hline
     0.344&0.227&0.250&0.314\\ \hline
     0.372&0.366&0.304&0.294\\ \hline
     0.075&0.090&0.072&0.077\\ \hline
     0.321&0.346&0.326&0.364\\
     0.363&0.379&0.351&0.400\\
     0.336&0.328&0.338&0.359\\ \hline
     0.228 &0.219&0.216&0.220\\ 
     0.326 &0.304&0.307&0.335\\ 
     0.295 &0.298&0.289&0.286\\ \hline
     0.200&0.176&0.152&0.173\\ \hline
     0.229&0.234&0.242&0.261\\
     0.285&0.263&0.299&0.298\\ \hline \hline
     20/30&22/30&22/30&21/30\\ \hline    
  \end{tabular}
  \caption{Classification error rates. Here, TN-C$_{Li}$ is the classifier learnt using embeddings from $i$th layer of Timenet. * means SAE from the first dataset in the group was used for training SAE-C.} 
  \label{tab:results}
\end{table}

Amongst the test datasets, there are four cases where multiple datasets belong to the same domain. In such cases, we train one SAE model per domain using the dataset with the largest number of train instances (refer Table \ref{tab:results}). However, the classifier SAE-C is trained on the respective training set.
We observe that TN-C and SAE-C exceed or match the performance of DTW-C on 83\% (25/30) datasets and 73\% (22/30) datasets, respectively. Also, TN-C exceeds the performance of SAE-C on 60\% (18/30) datasets. These results further confirm that \textit{a pre-trained Timenet gives embeddings which provide relevant features for TSC.}
TN-C is better than the recently proposed state-of-art PROP (proportional ensemble of classifiers based on elastic distance measures) on 4 out of 15 test datasets for which the results have been reported in \citep{lines2015time}.

\textbf{TN-C with reduced labeled training data}: We further test the robustness of embeddings by reducing the amount of labeled data used, and learn a classifier TN-C$_{2/3}$ using randomly selected two-thirds of the training data while maintaining class balance. We report average of the error rate of three such random selections in Table \ref{tab:results}. TN-C$_{2/3}$ performs better than DTW-C on 66\% (20/30) datasets suggesting that \textit{robust embeddings can be obtained using a pre-trained Timenet, and then a classifier can be learnt over the embeddings using lesser amount of labeled information.} 

\textbf{Performance of each layer of Timenet}: To evaluate the relevance of each layer of Timenet, we learn SVM classifier TN-C$_{Li}$ using embeddings from only the $i$th hidden layer. We observe that for datasets with small $T$, a single layer of Timenet gives reasonable classification performance, and at times performs better than TN-C. For datasets with large $T$, the classifier TN-C is better than any TN-C$_{Li}$ (refer Table \ref{tab:results}). This suggests that \textit{for shorter time series, one of the three layers extracts relevant features from time series, whereas for longer time series all layers carry relevant information.}

\section{Related Work}\label{sec:related}
Many approaches for TSC analyzed in \citep{lines2015time} use variations of time warping and edit distance in a nearest neighbor classifier. Other approaches extract statistical and frequency-based features from time series, and learn a classifier over these features.  To the best of our knowledge, our work is the first to show that it is possible to leverage unlabeled varying length time series from diverse domains to train a multilayered recurrent network as a feature extractor. 

Recurrent neural networks (RNNs) for TSC have been proposed in \citep{husken2003recurrent}. Several approaches for time series modeling using deep learning based on models such as RNNs, conditional Restricted Boltzmann Machines (RBM), temporal RBMs, Time-Delay Neural Networks, and Convolutional Auto-encoder have been surveyed in \citep{langkvist2014review}. \citep{langkvist2012sleep} proposed unsupervised feature learning using Deep Belief Networks (DBNs) for sleep stage classification from time series. The static models such as Convolutional auto-encoder and DBNs partition the time series into windowed time series of fixed length to analyze temporal data. Our approach proposes the seq2seq model which uses a pair of RNNs for learning representations without need of windowing, and may be more naturally suited for temporal data related tasks. Also, to the best of our knowledge, ours is the first approach which proposes a generic unsupervised pre-trained RNN as time series encoder for diverse univariate time series.

Very recently, \citep{oord2016wavenet} showed that it is possible to have one generic deep neural network for raw audio waveforms that can capture different characteristics of many different speakers with equal fidelty. \citep{chung2016unsupervised} proposed Audio Word2Vec model where they show usefulness of learning fixed-dimensional representations from varying length audio signals using seq2seq models. Our work extends the above approaches specific to speech domain, and shows that it is possible to learn a generic encoder for time series from diverse domains.
Semi-supervised sequence learning proposed in \citep{dai2015semi} shows that LSTMs pre-trained using sequence auto-encoders on unlabeled data are usually better at classification than LSTMs initialized randomly. This work uses the encoder network to initialize the LSTM classifier whereas we use the embeddings obtained from encoder for classification. Data augmentation techniques (e.g. \cite{yadav2015ode}, \cite{le2016data}) have been proposed to handle data scarcity for training deep models for sequences or time series. ODEs as a generative model for time series have been shown to improve performance of RNNs for anomaly detection \cite{yadav2015ode}. Convolutional Neural Networks for TSC \citep{le2016data} has been recently proposed where data augmentation through window slicing and warping is proposed to handle scarcity of data. Further, pre-training of each layer using auto-encoder is proposed using similar-length time series from other domains (data mixing) to improve performance. Our work is different from this approach in the sense that we use pre-trained encoder network learnt on diverse time series of varying length rather than leveraging slicing, warping, and data mixing from similar-length time series to handle scarcity of data for training deep networks.

\section{Discussion}\label{sec:conclusion}
Deep neural networks are data-intensive models. In practical applications, getting labeled data is costly although unlabeled data may be readily available. Recently, sequence-to-sequence models trained in an unsupervised manner as auto-encoders have been successfully demonstrated for spoken term detection from audio segments, and for anomaly detection and prognostics from sensor data. We exploit such a sequence auto-encoder trained on diverse unlabeled time series data to obtain a deep recurrent neural network (RNN) that transforms time series to fixed-dimensional representations or embeddings. This pre-trained multilayered RNN is found to yield effective embeddings for time series classification. The t-SNE visualizations further confirm the ability of Timenet to yield meaningful embeddings. The encoders of domain-specific sequence auto-encoders also provide useful representations for time series classification although they may be prone to overfitting given few training instances. Studying the usefulness of embeddings obtained from Timenet for other tasks such as anomaly detection and clustering is a plausible direction for future work.

\bibliography{bibTeX/phm-kdd,bibTeX/icml2016,bibTeX/nips2015,bibTeX/nips2015_2,bibTeX/esann,bibTeX/nips2016}

\appendix
\section*{Appendix A. Datasets used for training Timenet}
\begin{table}[h]
  \scriptsize
  \centering
  {\begin{tabular}{|cccc|cccc|} \hline
    \textbf{Dataset} & \textbf{T} & \textbf{C} &\textbf{N}&\textbf{Dataset}& \textbf{T} & \textbf{C}&\textbf{N}\\ \hline
    \textbf{ItalyPowerDemand} & 24 & 2&1096 &\textbf{SonyAIBORobotSurfaceII} & 65 & 2&980\\    
    \textbf{SonyAIBORobotSurface} & 70 & 2&621 & \textbf{TwoLeadECG} & 82 & 2&1162\\
    \textbf{FacesUCR} & 131 & 14 &2250& \textbf{Plane} & 144 & 7&210\\
    \textbf{Gun\_Point} & 150 & 2 &200& \textbf{ArrowHead} & 251 & 3&211\\    
     \textbf{WordSynonyms} & 270 & 25 & 905& \textbf{ToeSegmentation1} & 277 & 2&268\\
     \textbf{Lightning7} & 319 & 7 & 143&\textbf{ToeSegmentation2} & 343 & 2&166\\
     \textbf{DiatomSizeReduction} & 345 & 4 &322& \textbf{OSULeaf} & 427 & 6&442\\
     \textbf{Ham} & 431 & 2 &214& \textbf{Fish} & 463 & 7&350\\
     \textbf{ShapeletSim} & 500 & 2&200 & \textbf{ShapesAll} & 512 & 60&1200\\
     \hline
  \end{tabular}
  \subcaption{Training datasets}}
  \vspace{0.4em}
  \centering
  {\begin{tabular}{|P{3.75cm}ccc|P{3.99cm}ccc|} \hline      
     \textbf{Dataset} & \textbf{T} & \textbf{C}&\textbf{N} &\textbf{Dataset}& \textbf{T} & \textbf{C}&\textbf{N}\\ \hline
    \textbf{MoteStrain} & 84 & 2&1272 &\textbf{CBF} & 128 & 3&930 \\ 
    \textbf{Trace} & 275 & 4&200 & \textbf{Symbols} & 398 & 6 &1020\\ 
    \textbf{Herring} & 512 & 2&128 & \textbf{Earthquakes} & 512 & 2&461 \\\hline
  \end{tabular}
  \subcaption{Validation datasets}}
  \caption{Training and validation datasets used for Timenet. Here, T: time series length, C: number of classes, N: number of time series.} 
  \label{tab:datasets}
\end{table}
\end{document}